%% file: main.tex
\def\year{2021}\relax
\documentclass[letterpaper]{article} 
\usepackage{aaai21}  
\usepackage{times}  
\usepackage{helvet} 
\usepackage{courier}  
\usepackage[hyphens]{url}  
\usepackage{graphicx} 
\usepackage{graphicx}  
\usepackage{natbib}  
\usepackage{caption} 
\usepackage{tabularx}
\usepackage{amssymb}
\usepackage{enumerate}
\usepackage{amsfonts}
\usepackage{amsmath}
\usepackage[linesnumbered,commentsnumbered,ruled,vlined]{algorithm2e}
\usepackage{mathrsfs}
\usepackage{graphicx}
\usepackage{subfigure}

\title{Integrating Representation and Imitation in Reinforcement Learning for Mobile User Profiling: An Adversarial Training Perspective}

\title{Reinforced Imitative Graph Representation Learning for Mobile User Profiling: An Adversarial Training Perspective}

\author {
        Dongjie Wang\textsuperscript{\rm 1}
        Pengyang Wang \textsuperscript{\rm 1}
        Kunpeng Liu \textsuperscript{\rm 1}
        Yuanchun Zhou \textsuperscript{\rm 2}
        Charles Hughes \textsuperscript{\rm 1}
        Yanjie Fu \textsuperscript{\rm 1}\footnote{Contact Author}
       \\
}
\affiliations {
    \textsuperscript{\rm 1} University of Central Florida \\
    \textsuperscript{\rm 2} Computer Network Information Center, Chinese Academy of Sciences \\
    wangdongjie@knights.ucf.edu, pengyang.wang@knights.ucf.edu, kunpengliu@knights.ucf.edu, zyc@cnic.cn, charles.hughes@ucf.edu, yanjie.fu@ucf.edu
}

\begin{document}

\maketitle

\input{abstract}

\input{introduction}

\input{preliminaries}

\input{method}

\input{experiment}

\input{related}

\input{conclusion}

\newpage

\section*{Acknowledgment}
This research was partially supported by the National Science Foundation (NSF) via the grant numbers: 1755946, I2040950, 2006889.

\bibliography{aaai}


\end{document}

%% file: abstract.tex
\begin{abstract}
In this paper, we study the problem of mobile user profiling, which is a critical component for quantifying users' characteristics in the human mobility modeling pipeline.
Human mobility is a sequential decision-making process dependent on the users' dynamic interests. 
With accurate user profiles, the predictive model can perfectly reproduce users' mobility trajectories. 
In the reverse direction, once the predictive model can imitate users' mobility patterns, the learned user profiles are also optimal. 
Such intuition motivates us to propose an imitation-based mobile user profiling framework by exploiting reinforcement learning, in which the agent is trained to precisely imitate users' mobility patterns for optimal user profiles.
Specifically, the proposed framework includes two modules: (1) representation module, that produces state combining user profiles and spatio-temporal context in real-time;
(2) imitation module, where Deep Q-network (DQN) imitates the user behavior (action) based on the state that is produced by the representation module. 
However, there are two challenges in running the framework effectively. 
First, $\epsilon$-greedy strategy in DQN makes use of the exploration-exploitation trade-off by randomly pick actions with the $\epsilon$ probability. 
Such randomness feeds back to the representation module, causing the learned user profiles unstable. 
To solve the problem, we propose an adversarial training strategy to guarantee the robustness of the representation module.
Second, the representation module updates users' profiles in an incremental manner, requiring integrating the temporal effects of user profiles. 
Inspired by Long-short Term Memory (LSTM), we introduce a gated mechanism to incorporate new and old user characteristics into the user profile. 
In the experiment, we evaluate our proposed framework on real-world datasets. 
The extensive experimental results validate the superiority of our method comparing to baseline algorithms.

\end{abstract}

%% file: introduction.tex
\section{Introduction}
Mobile user profiling is to learn users' profiles from historical mobility records. 
Mobile user profiling has drawn significant attentions from various disciplines, such as mobile advertising~\cite{grewal2016mobile,haver2017systems}, recommended system~\cite{zhao2018stellar,yin2017spatial}, urban surveillance~\cite{yi2018integrated,wang2017non}, etc. 
In this paper, we study the problem of learning with a mixed user-event stream for mobile user profiling.

In recent literature, mobile user profiling is usually formulated as a representation learning task to map user behavior data into lower-dimensional vectors~\cite{wang2019adversarial,wang2020incremental}. 
To obtain optimal user profiles, we leverage a novel imitation-based learning criterion~\cite{wang2020incremental}, in which user profiles are optimal when an agent can perfectly imitate users' activities. 
In the imitation based criteria, we create an imitation task to mimic user activities over time based on the user profiles generated by the representation task, in order to provide performance feedback to improve user profiling.
Therefore, a joint and sequential learning paradigm is highly desired to achieve the mutual enhancement between the representation and imitation tasks.
The emerging reinforcement learning provides great potential to fill this gap: (1) reinforcement learning can model sequential decision making data, (e.g., mobile user visit sequences); 
(2) we can analogize user profiling as state learning, and imitation based on user profiles as policy learning. 
Along these lines, we propose a Reinforcement Imitative Representation Learning (RIRL) method for user profiling. 
Specifically, the proposed framework includes two modules:
(1) representation module, which generates state combing user profiles and spatio-temporal context in real-time, and 
(2) imitation module, which exploits an agent and a policy network to imitate the user behavior (action) based on the state that is generated by the representation module. 
The imitation performance will be evaluated as the feedback to the representation module for updating user profiles. 
The generated user profiles will be optimized once the imitation module can perfectly reproduce users' mobility trajectories. 


However, there are two challenges towards the practically deployment of the proposed framework. 
First, although the exploration in reinforcement learning improves imitated performance, it jeopardizes the quality of user profiles due to gradient sharing between imitation module and representation module.
Second, the representation module updates users' profiles in an incremental manner, requiring integrating the temporal effects of user profiles. 
We will detail how we tackle these two challenges as follows.

First, reinforcement learning aims at learning the optimal policy (imitation module) following the exploration-exploitation schema. 
For example, one representative strategy is $\epsilon$-greedy that randomly picks actions with probability of $\epsilon$ for exploring the action space. 
While the exploration benefits discovering better actions, the randomly picked actions (reproduced user mobility behavior) would result in biased imitation, which further jeopardize the quality of user profiles over time.
To guarantee robust user profiles, we develop a new problem setting that integrates reinforcement learning with the constraint of robust representation learning. 
We derive the solution for the new problem setting from an adversarial learning perspective.
Specifically, we regard the representation module as a generator, and the imitation module plus the reward function as the discriminator. 
The discriminator has two objectives: (1) imitating user behavior, which is updated based on reinforcement learning rule;
(2) providing a quality score for the user profile, which is determined by the reward function.
The scoring criteria of the discriminator is the reward function that produces values based on a certain rule (the definition of the reward function). 
Therefore, the discriminator has already achieved its optimal.
We only need to utilize the optimal discriminator to improve the performance of the generator.
We use the updating gradient, coming from the loss between the current reward and the maximum value of the reward, to update the parameters of the representation module.

Second, user profiles indicate the dynamic interests evolving over time. 
Therefore, the learned user profiles are expected to preserve such temporal effects. 
To capture such temporal effects as user profiles evolving, we propose a new updating schema. 
Intuitively, incrementally updated user profiles capture not only the most important old user features, but also identifies the newest user interest.
Such intuition suggests us to discard trivial and outdated information and merge new user-interests.
Inspired by Long-short Term Memory (LSTM), we introduce a gated mechanism into the profile updating process.
We first input an old user profile into a sigmoid neural unit to calculate the proportion of the old important information, which is denoted by a learnable parameter $\alpha$.
Then, we combine with the old and new information in the proportion of $\alpha$ and $1-\alpha$, respectively.
The more important the old information is, the larger $\alpha$ is.
In this way, the user profile preserves the most representative and important user characteristics.

In summary, we propose a new mobile user profiling framework that integrates the representation and imitation in reinforcement learning.
Our contributions are: (1) We develop a new reinforcement learning setting with the constraint of robust representation learning, and derive a solution from an adversarial perspective.
(2) We introduce a gated mechanism to make the user profile contain the most representative user characteristics by paying attention to the temporal effects.
(3) We conduct extensive experiments on two real-world check-in datasets to validate the effectiveness of our proposed framework.

%% file: preliminaries.tex
\section{Preliminaries}
\subsection{Definitions and Problem Statement}
In this section, we introduce the key definitions and problem statement of the user profiling framework (RIRL).

\begin{figure*}[!t]
    \centering
    \includegraphics[width=0.9\linewidth]{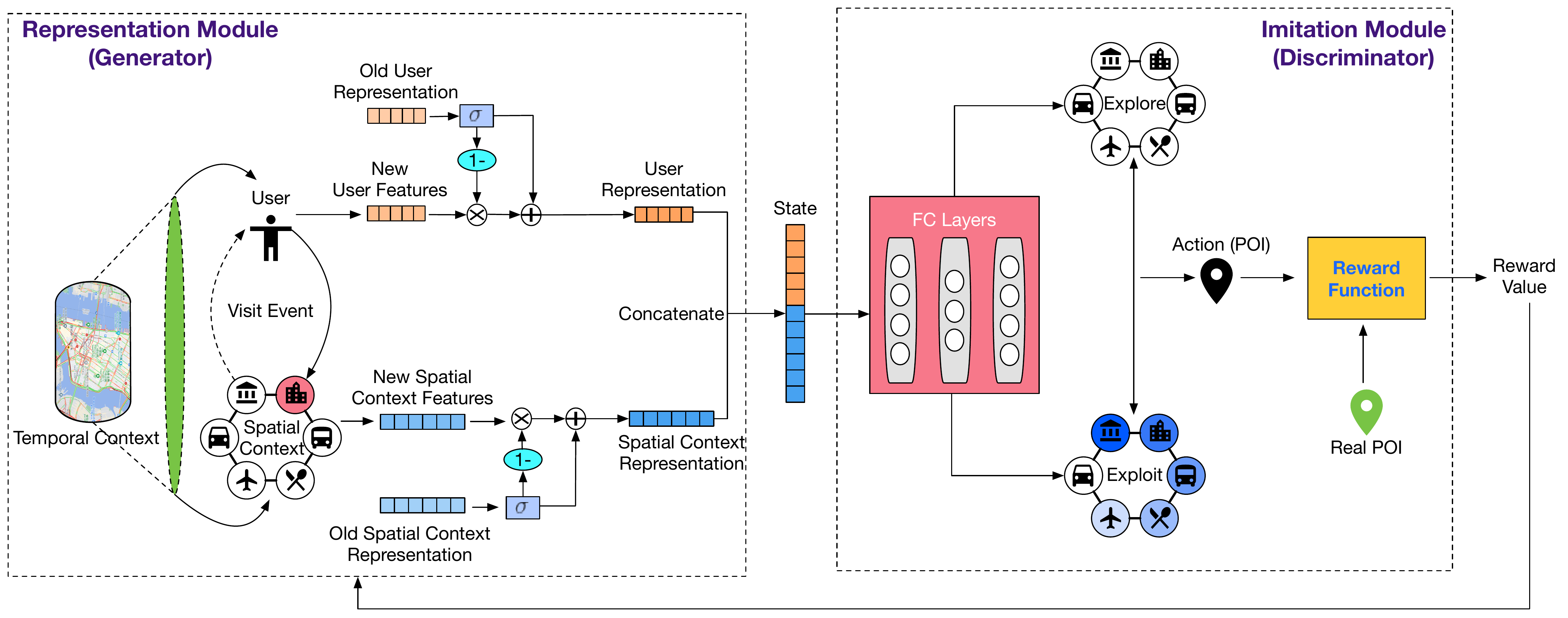}
    \vspace{-0.2cm}
    \captionsetup{justification=centering}
    \vspace{-0.15cm}
    \caption{An Example for RIRL framework.}
    \vspace{-0.55cm}
    \label{fig:framework overview}
\end{figure*}

\subsubsection{Key Components of RIRL}
Our RIRL framework formulates mobile user profiling into a reinforcement learning task.
Here we give the definitions and the key components in RIRL as follows:

\begin{enumerate}
    \item {\bf Agent.}
    We construct a next-visit model as the agent (Imitation Module).
    The agent predicts the next-visit POI for a given user based on the current state of the environment.
    
    \item{\bf Action.}
    For a given user, 
     action is the predicted POI by the imitation module, which represents the user's mobility decision.
     The action space is the number of POIs.
     Suppose the user will visit POI $P_i$ at one time step, it can be denoted as the user takes action $a_i$ at the time step.
     
    \item{\bf Environment.}
    RIRL is a model-based reinforcement learning framework.
    We build a model (representation module) to represent the environment.
    For a given user, the user travels in spatial context ({\it i.e.} POIs) over temporal context ({\it i.e.} traffic flow).
    Formally, the user interacts with spatial context by the event of visiting POIs.
    The interaction affects the representation of the user and spatial context bidirectionally.
    From the spatial context perspective, 
    the visit event, which changes the representation of spatial context.
     Inversely, from the user perspective, the changes in spatial context lead the user to make a corresponding decision for next-visit POI.
     The temporal context both affects the representation of spatial context and user, but it is independent, which changes over time.
    
    \item{\bf State.}
    The state is a snapshot of the environment at one time step.
    Formally, we denote the state at the step $l$ as $s^l=((\mathbf{u}^l,\mathbf{g}^l) | \mathbf{T}^l)$. Specifically,  $\mathbf{u}^l=\left\{\mathbf{u}^l_i | u_i \in \mathcal{U} \right \}$, where $\mathbf{u}_i^l$ denotes the representation (user profile) of the user $\mathbf{u}_i$, $\mathcal{U}$ is the collection of all users;
    $\mathbf{g}^l$ is the representation of the spatial context.
    $\mathbf{T}^l$ is the representation of the temporal context.
    The combination of $\mathbf{u}^l$,
    $\mathbf{g}^l$,
    and $\mathbf{T}^l$
    is regarded as the state at the step $l$.

    \item{\bf Reward.}
    In RIRL, if the user behavior is predicted more accurately, the reward value should be higher.
    Here, a POI visit event can be divided into three parts: POI location, POI category and certain POI, so we define the reward as the weighted sum of: (1)
    $r_d$, the reciprocal of the distance between the real and predicted visit POI;
    (2) $r_c$, the similarity between the real and predicted the category of visit POI.
    (3) $r_p$, whether the predicted visit POI is the real one.
     In addition, we introduce baselines into the reward to reduce the variance.
     Formally, we set a time window with size $n$ for $r_d$, $r_c$, $r_p$ respectively.
     Each time window moves forward for one step at each iteration.
     The mean value of each time window is regarded as the baseline of corresponding item.
     Here, we use
     $b_d, b_c, b_p$   
     to denote the baselines of $r_d,r_c,r_p$ respectively.
    The reward function can be represented as:
    \begin{equation}
        r = \sigma(\lambda_d \times (r_d-b_d) 
        + \lambda_c \times (r_c-b_c) + \lambda_p \times (r_p-b_p)),
    \end{equation}
    where $\lambda_d$ , $\lambda_c$ and $\lambda_p$ represent the weights for $r_d$, $r_c$ and $r_p$ respectively; 
    $\sigma$ denotes the sigmoid function that smooths the distribution of the reward.
    Intuitively, the reward function encourages the agent to search the user possible next-visit POIs under the areas that are close to the real visited POI rather than all POIs.
    Moreover, we utilize {\it GloVe}\footnote{\url{https://nlp.stanford.edu/projects/glove/}} pre-trained word embedding to replace the POI category name to calculate the cosine similarity between predicted and real POI categories as the similarity $r_c$.
\end{enumerate}

\subsubsection{Problem Statement}
     In this paper, we learn mobile user profiles incrementally within a mixed user visit-event by RIRL.
     Formally, given a mixed visit-event that contains users and spatio-temporal context, our purpose is to find a mapping function $f: (\mathbf{u}^{l}, \mathbf{g}^{l})|\mathbf{T}^l \rightarrow (\mathbf{u}^{l+1}, \mathbf{g}^{l+1})|\mathbf{T}^{l+1}$.
     The meaning of the function is that
     for a given temporal context $\mathbf{T}^l$ at the step $l$, 
     the function takes the combination between user and spatial context as input and outputs the new combination at the next step $l+1$.
     During the mapping process, the imitation module first predicts user next-visit POI based on the state at the step $l$.
     Then, the reward function takes the predicted and real POI as input to calculate the reward value.
     Next, the reward value is fed back $f$ to instruct the parameters update of $f$ at the step $l+1$.
     Finally, when the imitation module predicts user behavior perfectly, the user profile $\mathbf{u}$ is the most effective.

     

%% file: method.tex
\section{RIRL for Mobile User Profiling}

\subsection{An Example for RIRL}
Figure \ref{fig:framework overview} shows an application example of RIRL that includes two  modules: 
(1) representation module, which is used to produce the state of the environment. 
In the module, we preserve two kinds of effects into the state:
a) the mutual interaction between the users and spatio-temporal context;
b) the temporal effect between old and new user characteristics in profile.
(2) Imitation module, which makes a prediction for user's behavior and provides a quality score (reward value) for the prediction.
Formally, we develop a DQN as the imitation component in the module.
For a given user, the DQN takes the state of the environment as input and integrates the exploration and exploitation to 
predict the next-visit POI (action) of the user.
Then the predicted POI and real POI are input into reward function to identify the performance of the prediction.
Next, We regard the reward value as a feedback of the imitation module to instruct the parameters update of the representation module.
The representation module and imitation module promote each other by the reward value.
When the imitation module mimics the user behavior perfectly,
the closed-loop system  produces the most effective user profile.



\subsection{Imitation Module}
RIRL is an imitation-based user profiling framework, therefore,
we first introduce the imitation module, which duplicates user behavior based on the state of the environment.
Deep Q-Network ({\it DQN})\cite{mnih2013playing} is a classical deep reinforcement learning model that learns the policy based on the state of the environment.
Here, the imitation module is implemented by {\it DQN}. 
To be convenient, we take the state $s^l$ at step $l$ as an example to explain the whole process.
Formally, the imitation module first takes the state $s^l=(\mathbf{u}^l,\mathbf{g}^l)$ as input.
Then, it
utilizes $\epsilon-greedy$ strategy to integrate exploration and exploitation for finding the most possible POI.
Specifically, the imitation module chooses a random POI $a^l_r$ 
with probability $\epsilon$ or POI $a^l_m$ that owns the maximum $Q$ value as prediction with probability $1-\epsilon$, the operation can be denoted by $a^l_m = \underset{a}{argmax}(Q(s^l,a^l))$. 
Next, the predicted POI and real POI are input into the reward function to calculate the reward value, and the imitation module utilizes the reward value to update parameters based on the Bellman Equation.
The loop continues until the module copies the user behavior perfectly.

In order to utilize the user trajectory data efficiently and break the strong correlation between data samples, we exploit the prioritized experience replay strategy to accelerate the learning procedure of {\it DQN} \cite{schaul2015prioritized}.
There are two steps in the strategy: (1) we first assign priority score for each data sample $(s^l,a^l,r^l,s^{l+1})$; (2) we then construct a priority distribution based on the priority score for sampling batch of data from memory. 
In RIRL, we adopt two sampling strategies from~\cite{wang2020incremental}:
(1) Reward-based, which sets the reward value of the corresponding data sample as the priority score. 
For a given data sample,
the higher reward of the sample is, the larger possibility is to be sampled.
(2) Temporal Difference (TD)-based, which sets the corresponding TD-error as the priority score.
For a given data sample, the larger TD-error is, the sample contains more valuable information to be learned.
The l-th step TD-error can be represented as:
\begin{equation}
\vspace{-0.1cm}
    \text{TD-error}^l=r^{l}+\gamma \max_{a^{l+1}}Q(s^{l+1}, a^{l+1}) - Q(s^{l}, a^{l}),
    \vspace{-0.1cm}
\end{equation}
After assigning the priority score for each data sample, we calculate the distribution of the priority score.
Specifically, a softmax function takes the priority score as input and outputs the priority distribution.
Then we sample top $k$ data based on the priority distribution as a batch to train the imitation module, where $k$ is the size of one batch.

\subsection{ Representation Module}
In this section, we show the representation module, which generates the state of the environment in real-time.
Representation module contains two components:
(1) modeling interaction between users and spatio-temporal context;
(2) modeling temporal dependency of user profiles.
For the first component,
the spatial context is a spatial knowledge graph ({\it KG}) that indicates the geographical environment.
We denote the spatial {\it KG} as
$\mathbf{g}^l = < \mathbf{h}^l, \mathbf{rel}, \mathbf{t}^l >$,
where $\mathbf{h}^l$ is the representation of the heads ({\it i.e.}, POIs), $\mathbf{t}^l$ is the representation of the tails ({\it i.e.}, categories and functional zones),
$\mathbf{rel}$ is the representation of the relationship between the heads and tails.
The temporal context $\mathbf{T} \in \mathbb{R}^{M \times 3}$ is the combination of inner traffic, in-flow traffic, and out-flow traffic in all areas of a city, $M$ represents the number of areas, and $3$ represents the three kinds of traffic flows.
The interaction between user and spatial {\it KG} occurs accompanying the temporal context.
The users' visit events change the representation of spatial {\it KG}.
Inversely, a new spatial {\it KG} representation stimulates the users to choose next-visit place.
For the second component, new user profile should not only captures the newest user interest changes but also grasps the old representative user characteristics.
Without the loss of the generality, we take the POI visit event that a user $u_i$ visits the POI $P_j$ at step $l$ as an example to explain the whole process, the user profile will be updated for the step $l+1$.

{\bf User}.
We incorporate the user state $\mathbf{u}_i^l$ and the interaction between the POI $P_j$ and the user $u_i$ into $\mathbf{u}_i^{l+1}$ such that:
\begin{equation}
    \mathbf{u}_i^{l+1} = \sigma(\alpha_u \times \mathbf{u}_i^{l} + 
    (1-\alpha_u) \times (\mathbf{W}_u \cdot (\mathbf{h}_{P_j}^{l})^{\intercal} \cdot  \tilde{\mathbf{T}}^l)),
\end{equation}
where
$\mathbf{W}_u \in \mathbb{R}^{N \times 1}$ is the weight; $\alpha_u$ is a scalar that represents the proportion of old profiling information in $\mathbf{u}_i^{l+1}$ , it is calculated by:
\begin{equation}
    \alpha_u = \sigma(\mathbf{W}_{\alpha_u} \cdot \mathbf{u}_i^l + \mathbf{b}_{\alpha_u}),
\end{equation}
where $\mathbf{W}_{\alpha_u} \in \mathbb{R}^{1 \times N}$ is the weight and $\mathbf{b}_{\alpha_u} \in \mathbb{R}^{1 \times 1}$ is the bias term; 
$\tilde{\mathbf{T}}^{l} \in \mathbb{R}^{N\times1}$ is the temporal context vector adaptable with state update, it can be calculated by:
\begin{equation}
    \tilde{\mathbf{T}}^l = \sigma(\mathbf{W}_{T_1} \cdot \mathbf{T}^{l} \cdot \mathbf{W}_{T_2} + \mathbf{b}_T),
\end{equation}
where  $\mathbf{W}_{T_1} \in \mathbb{R}^{N\times M}$, $\mathbf{W}_{T_2} \in \mathbb{R}^{3\times1}$ and $\mathbf{b}_T \in \mathbb{R}^{N\times1}$ are the weights and bias respectively.


{\bf Spatial KG}.
During the updating process of the spatial {\it KG}, we only focus on directly visited POI $P_j$ and other POIs $P_{j^-}$ that ``belong to" the same category or ``locate at" the same functional zones with the directly visited $P_j$.
Here, heads are POIs and tails are categories or functional zones.
Formally, we need to update the spatial {\it KG} representation $\mathbf{g}^l = < \mathbf{h}^l, \mathbf{rel}, \mathbf{t}^l >$.
We update the information in  $\mathbf{h}^l$ and $\mathbf{t}^l$, except  $\mathbf{rel}^l$.
In addition, $\sigma{(\cdot)}$ denotes the sigmoid function in following formulas.

\begin{enumerate}[(1)]
    \item Updating visited POI $\mathbf{h}_{P_j}^{l+1}$
    
    Similar to update  $\mathbf{u}^{l+1}_i$, we incorporate the old visited POI state $\mathbf{h}^{l}_{P_j}$ and the interaction between the use $\mathbf{u}_i^l$ and the POI $P_j$ in a given temporal context:
    \begin{equation}
           \mathbf{h}_{P_j}^{l+1} = \sigma(\alpha_p \times \mathbf{h}_{P_j}^{l} + (1-\alpha_p) \times (\mathbf{W}_p \cdot (\mathbf{u}_{i}^{l})^{\intercal} \cdot  \tilde{\mathbf{T}}^l)),
           \vspace{-0.1cm}
    \end{equation}
    where $\mathbf{W}_p \in \mathbb{R}^{N \times 1}$ is the weight;
    $\alpha_p$ is a scalar that denotes the proportion of old POI information in $\mathbf{h}_{P_j}^{l+1}$, it is calculated by:
    \begin{equation}
        \alpha_p = \sigma(\mathbf{W}_{\alpha_p} \cdot \mathbf{h}_{P_j}^l +\mathbf{b}_{\alpha_p}),
    \end{equation}
    where $\mathbf{W}_{\alpha_p} \in \mathbb{R}^{1 \times N}$ is weight and $\mathbf{b}_{\alpha_p} \in \mathbb{R}^{1 \times 1}$ is bias.

    \item Updating category and functional zones (tail) $\mathbf{t}_{P_j}^{l+1}$
    
    We update tail $\mathbf{t}_{P_j}^{l+1}$ by the combination of $\mathbf{t}_{P_j}^{l}$, $\mathbf{h}_{P_j}^{l+1}$ and $\mathbf{rel}_{P_j}$, it can be denoted as:
    {\small
    \begin{equation}
    \mathbf{t}_{P_j}^{l+1} =
    \alpha_t \times \mathbf{t}_{P_j}^{l} +
    (1-\alpha_t) \times (\mathbf{h}_{P_j}^{l+1} + \mathbf{rel}_{P_j}).
    \end{equation}
    }
    where $\alpha_t$ is scalar that denotes the proportion of old tail information in $\mathbf{t}_{P_j}^{l+1}$, it is calculated by:
    \begin{equation}
    \alpha_t = \sigma(\mathbf{W}_{\alpha_t} \cdot \mathbf{t}_{P_j}^{l} +\mathbf{b}_{\alpha_t}),
    \end{equation}
    where $\mathbf{W}_{\alpha_t} \in \mathbb{R}^{1 \times N}$ is weight and $\mathbf{b}_{\alpha_t} \in \mathbb{R}^{1 \times 1}$ is bias. 
    
    \item Updating same category and location POIs $\mathbf{h}^{l+1}_{P_{j-}}$
    
    We update the other POIs that belong to the same POI category and locate at the same functional zones of the visited $P_j$ as
    \small
    \begin{equation}
    \left\{
             \begin{array}{lr}
             \mathbf{h}^{h}_{P_{j^-}} =\mathbf{t}_{P_{j^-}}^{l+1} -  \mathbf{rel}_{P_{j^-}}, &  \\
         \mathbf{h}^{l+1}_{P_{j^-}}=
             \sigma[\alpha_{h} \times \mathbf{h}^{l}_{P_{j^-}} + (1-\alpha_{h}) \times \mathbf{h}^{h}_{P_{j^-}} ]
             \end{array}
    \right.
    \end{equation}
    where $\alpha_h$ is a scalar that is the proportion of  $\mathbf{h}^{l}_{P_{j^-}}$ in $\mathbf{h}^{l+1}_{P_{j^-}}$, it is calculated by:
    \begin{equation}
    \alpha_h = \sigma(\mathbf{W}_{\alpha_h} \cdot \mathbf{h}_{P_{j^-}}^{l} +\mathbf{b}_{\alpha_h}),
    \end{equation}
    where $\mathbf{W}_{\alpha_h} \in \mathbb{R}^{1 \times N}$ is weight and $\mathbf{b}_{\alpha_h} \in \mathbb{R}^{1\times 1}$ is bias.
\end{enumerate}

\subsection{ Application: Extracting User Profiles}
After the representation module finishes the above calculations at step $l$, the user representation (user profile) $\mathbf{u}^{l+1}$ and the spatial {\it KG} representation $\mathbf{g}^{l+1}$ are concatenated to construct the state  $s^{l+1}$ at step $l+1$.
Then $s^{l+1}$ is input into the imitation module to predict next-visit POI of the user.
The loop continues until the imitation module duplicate user mobility ideally based on the most effective user profile.

\section{Adversarial Training }
RIRL is a closed-loop learning system.
Since the imitation module is implemented by {\it DQN}, it brings randomness for the whole system.
The randomness is beneficial for the imitation module to integrate exploitation and exploration for predicting user behavior, but it may cause the learning procedure of the representation module is unstable and sensitive.
In order to solve this problem, we propose a new training strategy based on the adversarial learning perspective.

First of all, we interpret RIRL from an adversarial learning perspective. 
we regard the representation module as a generator to produce the state  in real-time.
Then, we treat the imitation module and reward function as a discriminator.
When the discriminator provides the maximum quality score continually, the representation module achieves the best situation.
Different from classical adversarial learning, the scoring criteria of RIRL is determined.
This is because, in the discriminator, the identifying criteria is a rule-based reward function, which indicates the discriminator has achieved its optimal.
Therefore, we adopt the most advanced discriminator to instruct parameters update in the generator.
Specifically, we utilize the gradient that comes from the gap between current reward and the expected reward value to update the parameters in the representation module.
Algorithm \ref{alg:feedback} shows the training process.

\vspace{-0.3cm}
\begin{algorithm}[htp] 
  \textbf{Denotation:}
  \newline$r$: reward function;\newline
  $\mathcal{I}$: imitation module;\newline
  $\mathcal{R}$: representation module ;\newline
  $a$: Real action (real POI).
  
  \For{number of training iterations}{
    Assume the loop variable is $i$.
    
    Update $\theta_\mathcal{R}$ by descending the gradient:\\
    \vspace{0.1cm}
    \begin{center}
         $\bigtriangledown_{\theta_\mathcal{R}}  log(1-r(\mathcal{I}(
     \mathcal{R}^i),a^i)).$
    \end{center}
  }
  \caption{Stochastic Gradient Descent Training for the Representation Module.}
  \label{alg:feedback}
\end{algorithm}
\vspace{-0.5cm}

Without the loss of generality, we use the $i$-th training iteration to explain Algorithm \ref{alg:feedback}.
We first generate the state at the step $i$ by $\mathcal{R}^i$.
Then, the imitation module $\mathcal{I}$ takes the state as input and outputs the predicted action (POI). 
Next, we use the reward function $r$ to calculate reward value based on predicted POI and real-visit POI.
Finally, we exploit the gradient that comes from the gap between $1$ ({\it max reward value}) and the current reward to update the parameters $\theta_\mathcal{R}$.

%% file: experiment.tex
\section{Experimental Results}

We conduct extensive experiments to answer the following research questions:
\begin{enumerate}[(1)]
    \item Is it effective to integrate representation and imitation in reinforcement learning for user profiling (RIRL)?
    \item Can the adversarial training and gated mechanism improve the model performance?
    \item Is our user profiling framework (RIRL) robust?
\end{enumerate}


\subsection{Data Description}
We conduct experiments on two check-in datasets from two cities: New York \cite{yang2014modeling} and Beijing \cite{wang2018ensemble}.
The characteristics of each dataset include: User ID, Venue ID, Venue Category ID, Venue Category Name, Latitude, Longitude, and Time.
Table \ref{table:data_stat_nyc} shows the statistics of the two datasets.

\begin{table}[t!hbp]
	\vspace{-0.25cm}
	\centering
	\scriptsize
	\tabcolsep 0.04in
	\caption {Statistics of the checkin data.}
		\vspace{-0.3cm}
	\begin{tabular}[t]{c|c|c|c|c}
		\hline
		\textbf{City}         & \textbf{\# Check-ins} & \textbf{\# POIs} &\textbf{\# POI Categories} & \textbf{Time Period}  \\ \hline
		New York & $227,428$ & $38,334$ &251 & 4/2012-2/2013 \\ \hline
		Beijing & $6,465$ & $3,434$ &9 & 3/2011-5/2011 \\
		\hline
	\end{tabular}
 	\vspace{-0.4cm}
	\label{table:data_stat_nyc}
\end{table}

In addition, we collect the taxi order data in New York and Beijing for calculating the temporal context $\mathbf{T}$.
Each taxi order includes ID, Pick-up: Longitude, Latitude, Time, and Drop-off: Latitude,  Longitude, Time.
We set the time window as one-hour to split the taxi order data.
\begin{table}[t!hbp]
 	\vspace{-0.35cm}
	\centering
	\scriptsize
	\tabcolsep 0.04in
	\caption {Statistics of the taxi data.}
		\vspace{-0.4cm}
	\begin{tabular}[t]{c|c|c }
		\hline
		\textbf{City}         & \textbf{\# Transactions} & \textbf{Time Period}  \\ \hline
		New York & $161,211,550$ & 4/2012-2/2013 \\ \hline
		Beijing & $12,000,000$ & 3/2011-5/2011 \\
		\hline
	\end{tabular}
 	\vspace{-0.55cm}
	\label{table:data_stat_bj}
\end{table}

\begin{figure*}[!thb]
	\centering
	\subfigure[Precision on Category]{\label{fig:substructure_precision_newyork}\includegraphics[width=4.35cm]{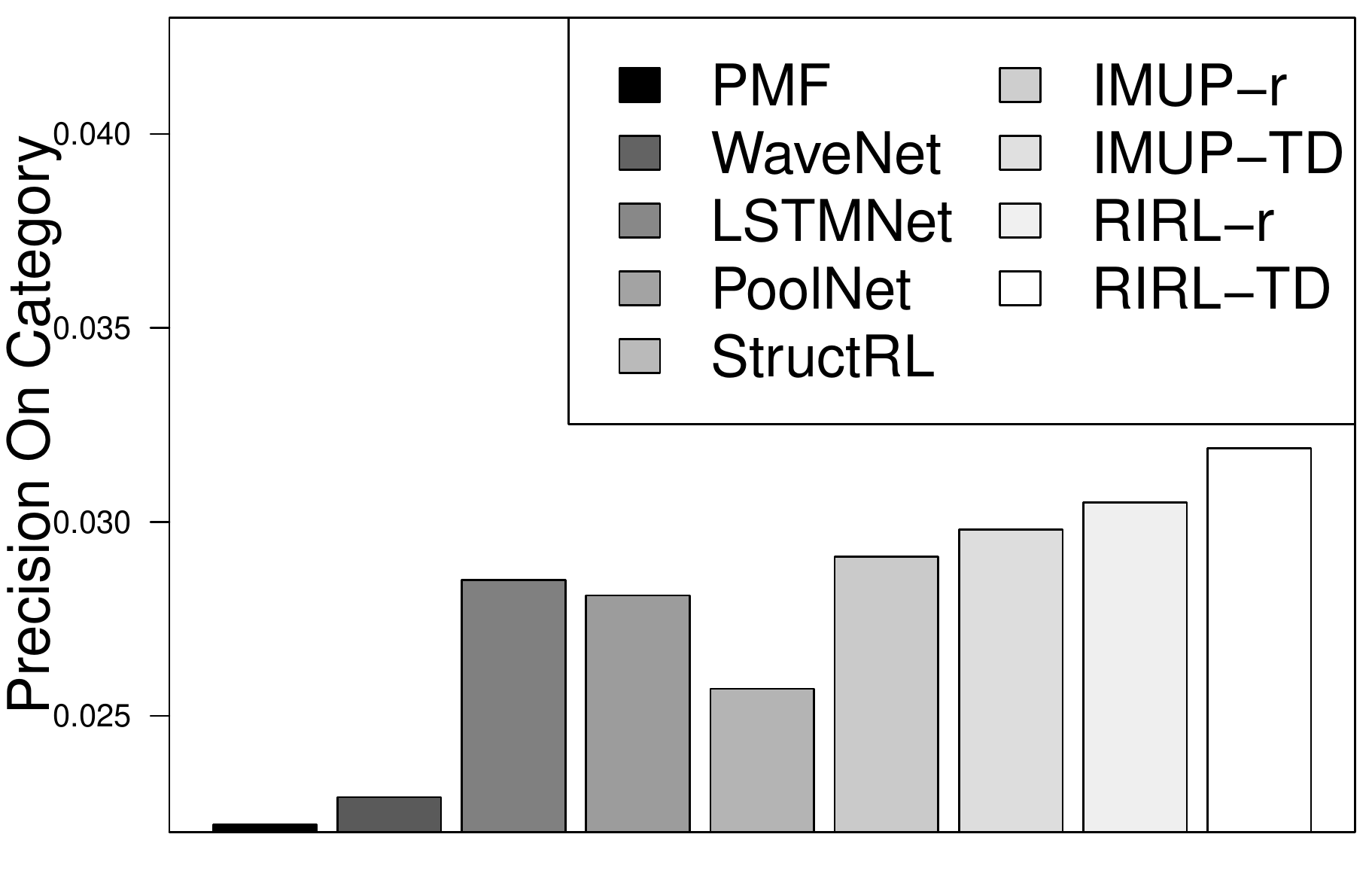}}
	\subfigure[Recall on Category]{\label{fig:substructure_newprecision_newyork}\includegraphics[width=4.35cm]{{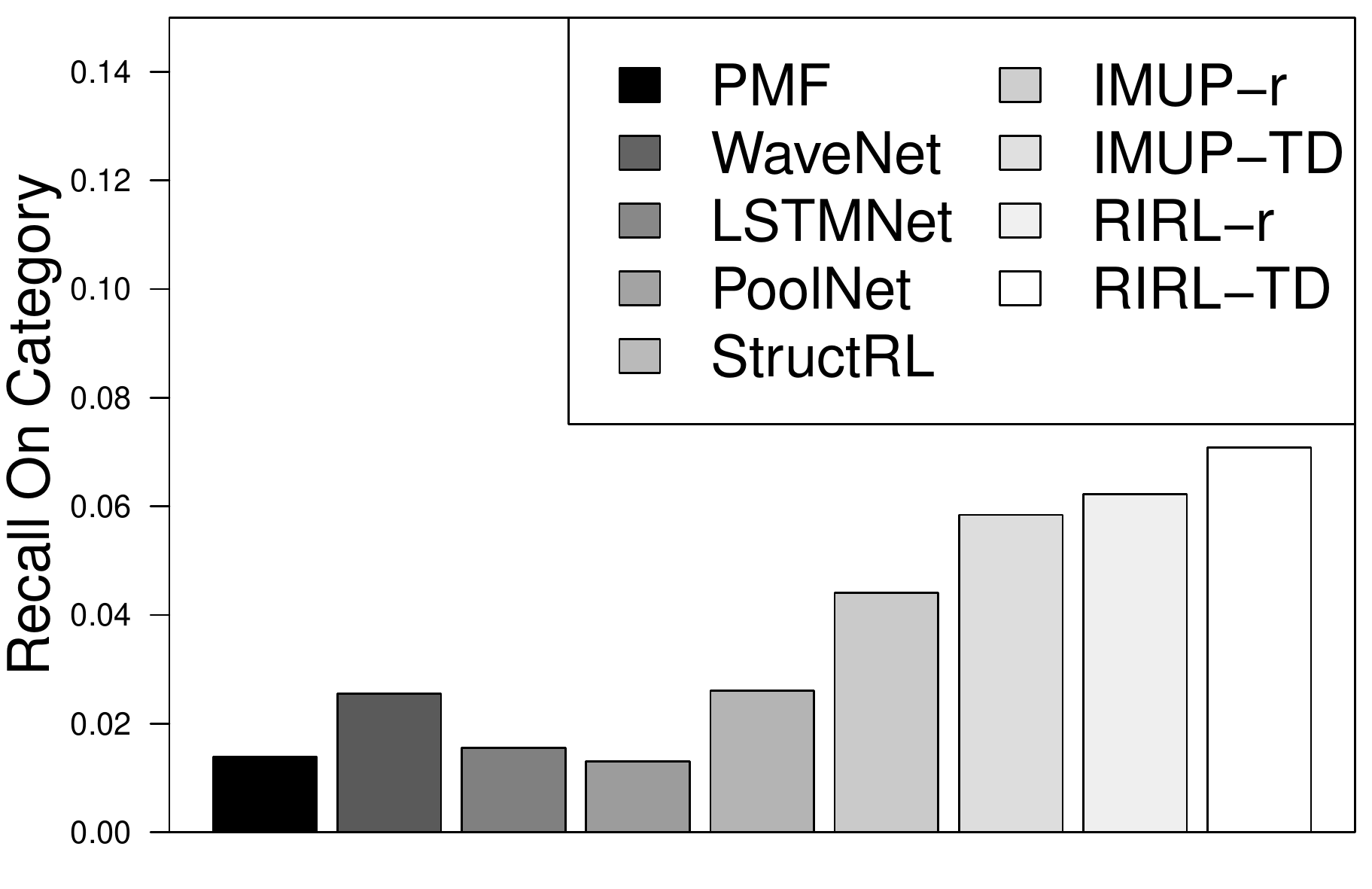}}}
	\subfigure[Average Similarity]{\label{fig:substructure_precision_tokyo}\includegraphics[width=4.35cm]{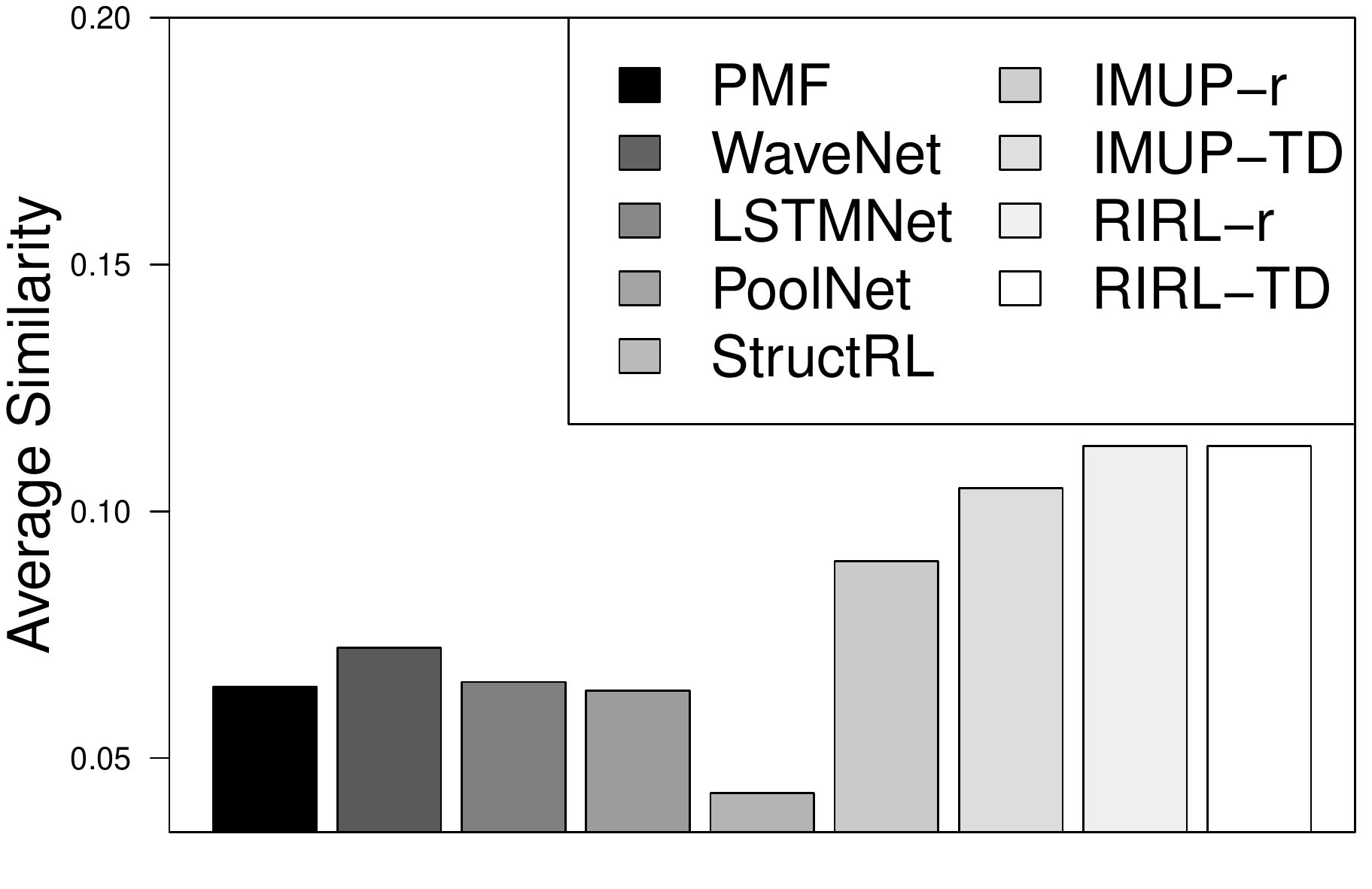}}
	\subfigure[Average Distance]{\label{fig:substructure_newprecision_tokyo}\includegraphics[width=4.35cm]{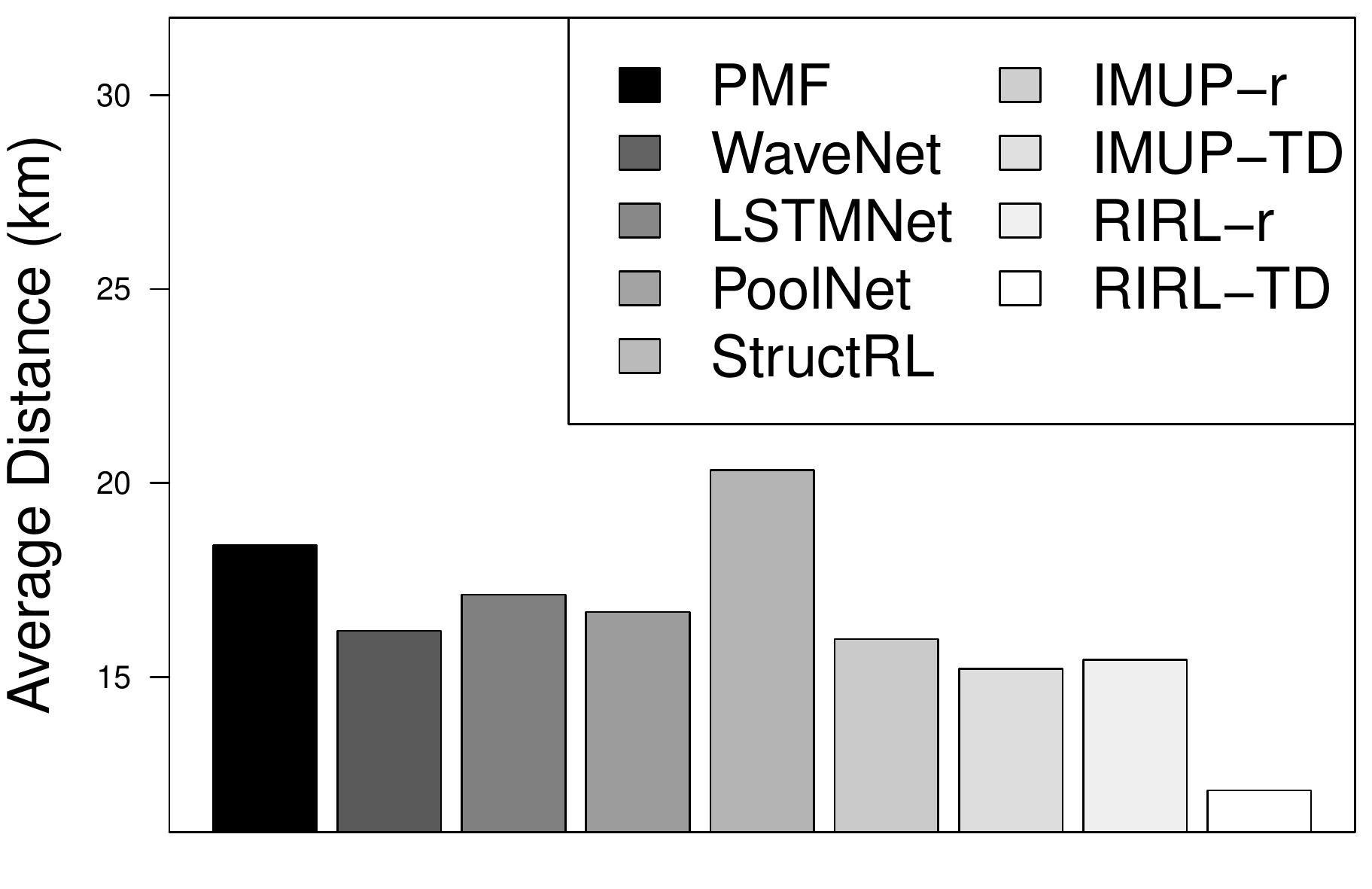}}
		\vspace{-0.4cm}
	\captionsetup{justification=centering}
	\caption{Overall comparison {\it w.r.t.} New York dataset.}
		\vspace{-0.3cm}
	\label{fig:nyc_overall}
\end{figure*}

\begin{figure*}[!tb]
	\centering
	\subfigure[Precision on Category]{\label{fig:substructure_precision_newyork}\includegraphics[width=4.35cm]{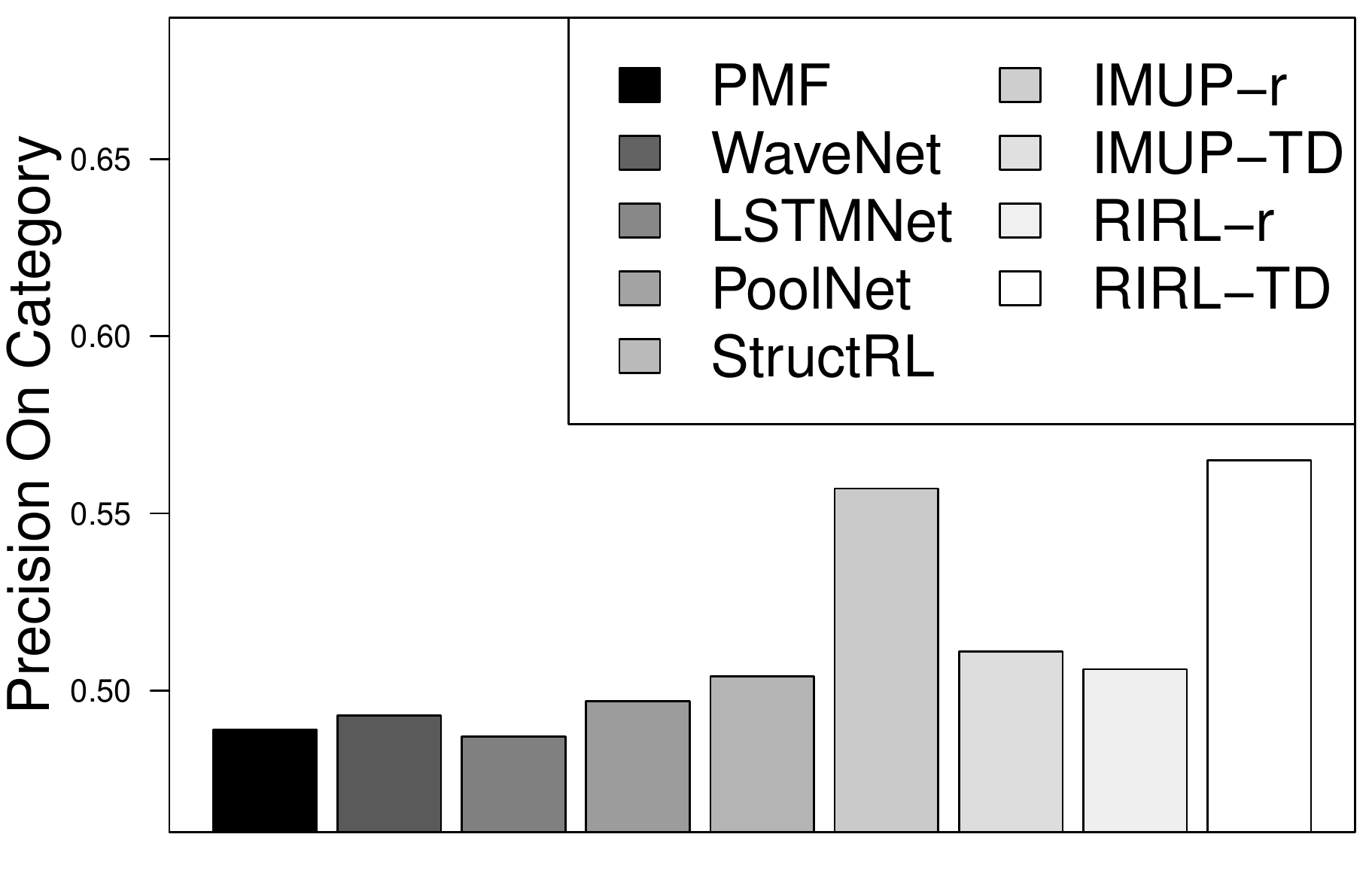}}
	\subfigure[Recall on Category]{\label{fig:substructure_newprecision_newyork}\includegraphics[width=4.35cm]{{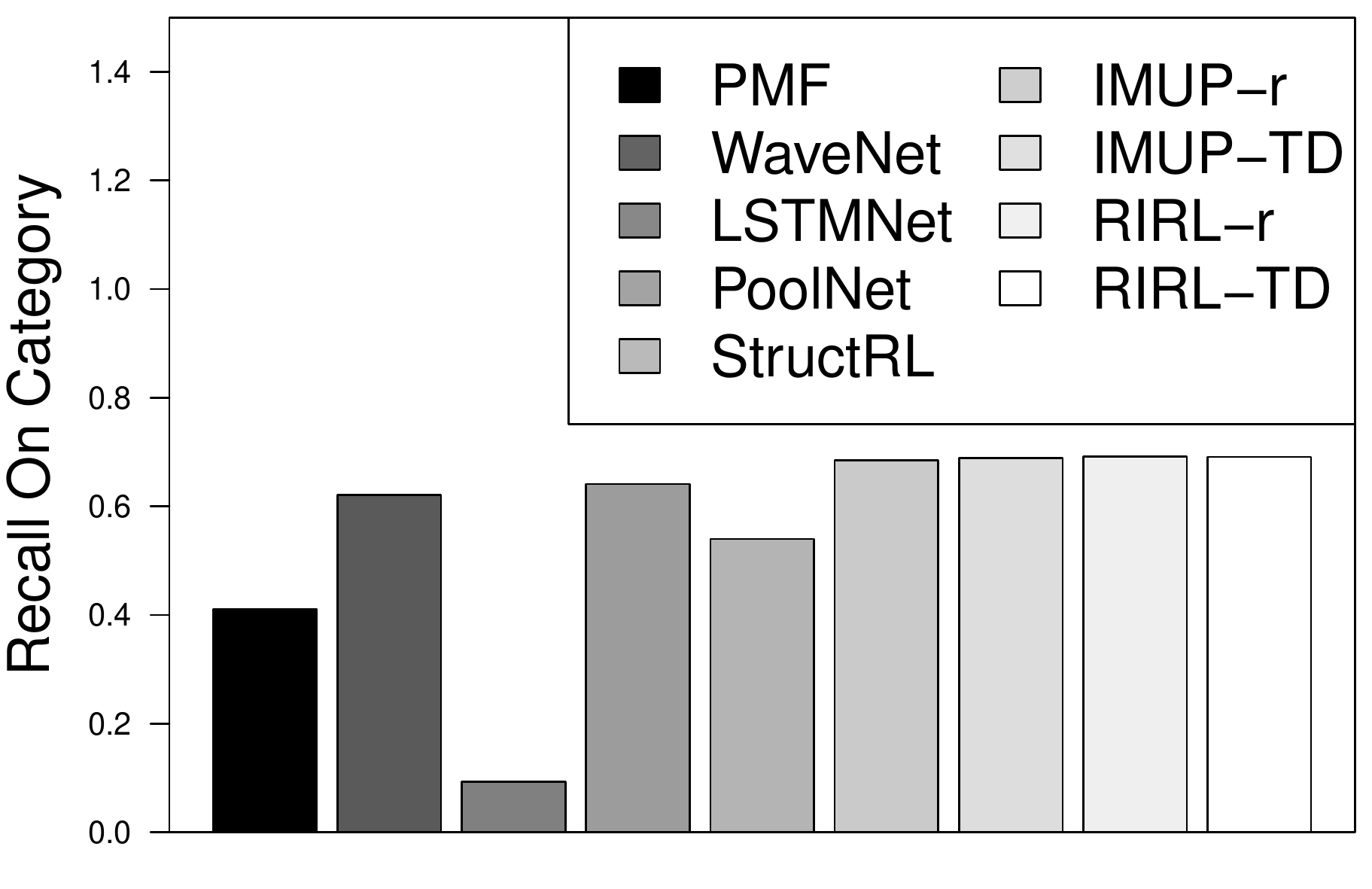}}}
	\subfigure[Average Similarity]{\label{fig:substructure_precision_tokyo}\includegraphics[width=4.35cm]{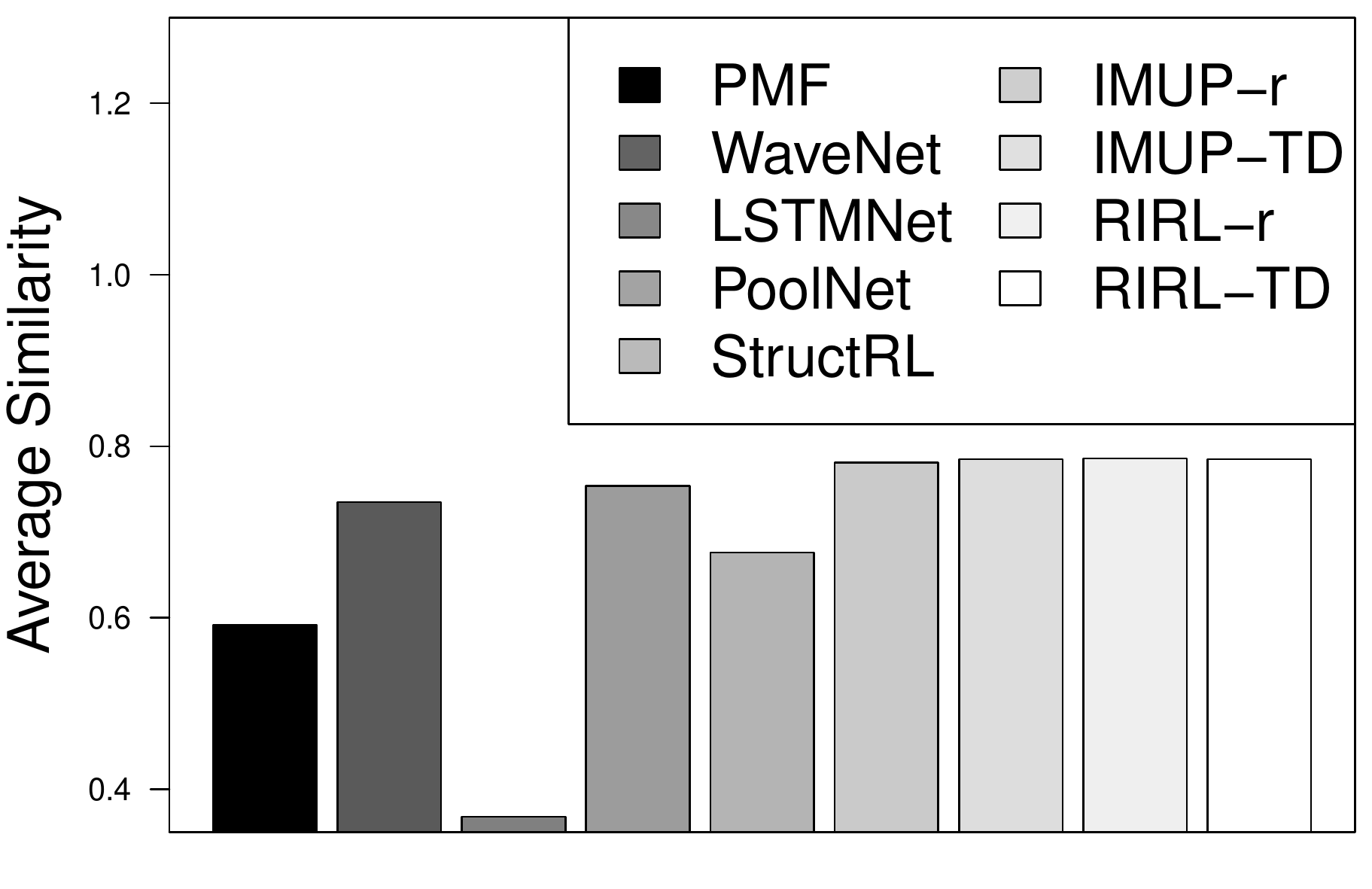}}
	\subfigure[Average Distance]{\label{fig:substructure_newprecision_tokyo}\includegraphics[width=4.35cm]{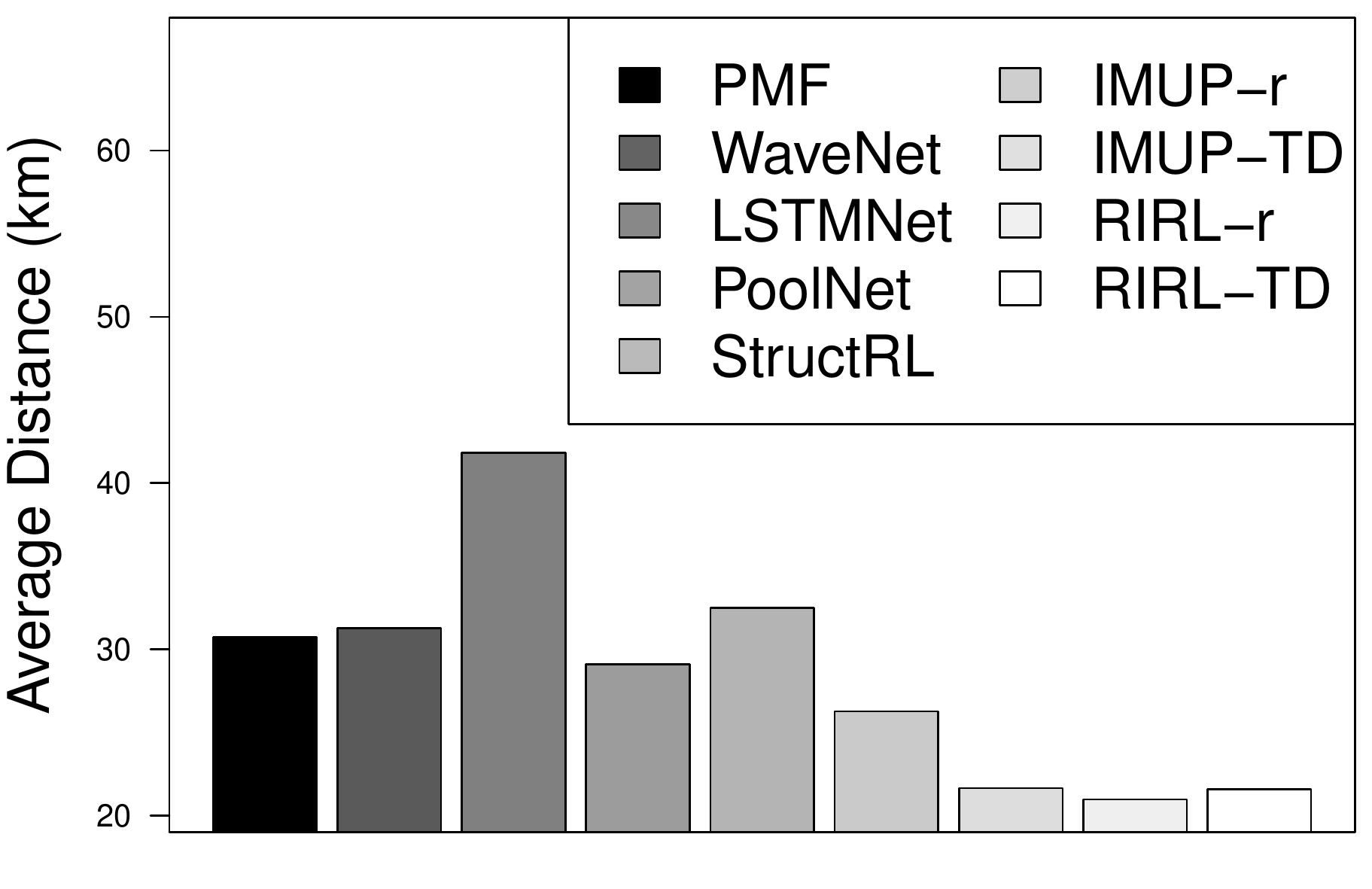}}
		\vspace{-0.4cm}
	\captionsetup{justification=centering}
	\caption{Overall comparison {\it w.r.t.} Beijing dataset.}
		\vspace{-0.6cm}
	\label{fig:bj_overall}
\end{figure*}

\subsection{Evaluation Metrics}
We assess the model performances over the predicted action (POI) on the POI categories and locations in terms of the following four metrics:
\begin{enumerate}[(1)]
    \item \textbf{ Precision on Category (Prec\_Cat):}
    We utilize the weighted precision to evaluate the classification precision on POI category.
    Here, $c_k$ represents $k$-th POI category; 
    $|c_k|$ is the number of $c_k$; $I_{TP}^k$ denotes the number of true positive predictions, and $I_{FP}^k$ denotes the number of false positive prediction.
    The formula of Prec\_Cat can be denoted as 
    \begin{equation}
        Prec\_Cat = \frac{|c_k| \cdot I_{TP}^k}{\sum \limits_k |c_k| (I_{TP}^k+I_{FP}^k)}
    \end{equation}
    
    \item \textbf{Recall on Category (Rec\_Cat):} 
    Following the definition of Prec\_Cat,
    we use the weighted recall to evaluate the recall on POI category.
    Here  $I_{FN}^{k}$ denotes the false negative prediction for $c_k$, Rec\_Cat can be denoted as 
    \begin{equation}
        Rec\_Cat = \frac{|c_k| \cdot I_{TP}^k}{\sum \limits_k |c_k| (I_{TP}^k+I_{FN}^k)}
    \end{equation}
    
    \item \textbf{Average Similarity (Avg\_Sim):}
    We evaluate the average similarity between the predicted and real POI categories.
    Here, ``$type^l$'' represents the word vector of real POI category; "$\hat{type^l}$" represents the word vector of predicted POI category; 
    $L$ denotes the total visit number, 
    Avg\_Sim can be denoted as 
    \begin{equation}
    Avg\_Sim = \frac{\sum \limits_{l} cosine(type^l, \hat{type^l})}{L}.
    \end{equation}

    The higher Avg\_Sim value represents the predicted POI category  is more similar to the real POI category.
    
    \item \textbf{Average Distance (Avg\_Dist):} 
    We utilize the average distance to evaluate the prediction on location. 
    Here, $Dist(P^l,\hat{P}^l)$ denotes the distance 
    between the real POI location $P^l$ and the predicted POI location $\hat{P}^l$ for the $l$-th visit, then the average distance is 
    \begin{equation}
    Avg\_Dist = \frac{\sum \limits_{l} Dist(P^{l}, \hat{P}^{l})}{L}.
    \end{equation}
    The lower Avg\_Dist value is, the closer between the predicted POI location and the real POI location is. 
\end{enumerate}

\subsection{Baseline Algorithms}

We compare the performances of our method RIRL against the following baseline algorithms:

\begin{enumerate}[(1)]
    \item \textbf{PMF.}
    Probabilistic matrix factorization (PMF) utilizes the probabilistic matrix factorization for producing user profiling 
 \cite{mnih2008probabilistic}.
    
    \item \textbf{PoolNet.}
    PoolNet produces the user representations by averaging the representations of different interacted items  \cite{covington2016deep}.
    
    \item \textbf{WaveNet.}
    WaveNet uses stacked causal atrous convolutions
    to  
    simulate sequential decision-making for
    learn user representation
     ~\cite{oord2016wavenet}.
    
    \item \textbf{LSTMNet.}
    LSTMNet preserves hidden user patterns in visiting sequence as user profile
     ~\cite{hidasi2015session}.
    
    \item \textbf{StructRL.}
    StructRL learns user profile within the adversarial learning paradigm 
  ~\cite{wang2019adversarial}.
    
    \item \textbf{IMUP-r.}
   IMUP integrates reinforcement learning and spatial {\it KG} to solve user profiling problem.
   IMUP-r is that the reinforced model in IMUP utilizes reward-based sampling strategy.
 ~\cite{wang2020incremental}.
    
    \item \textbf{IMUP-TD.}
     The problem formulation is the same as IMUP-r.
    The difference is that the sampling strategy is TD-based in the model.
    \cite{wang2020incremental}.
\end{enumerate}

Moreover, our proposed method RIRL has two variants: (1) \textbf{RIRL-r}, where the imitation model uses the reward-based sampling strategy.
(2) \textbf{RIRL-TD}, where the imitation model uses TD-based sampling strategy.

For the details of experiment settings, please refer to the appendix.

\subsection{Overall Comparison}

Figure \ref{fig:nyc_overall} and Figure \ref{fig:bj_overall} present that our ``RIRL-r'' and ``RIRL-TD'' outperform other baseline models in terms of ``Precision on Category'', ``Recall on Category'', ``Average Distance'' and ``Average Similarity'' over both the New York and Beijing dataset.
The result validates the effectiveness of the user profiling framework that integrates the representation and imitation in reinforcement learning.
\begin{figure*}[!thb]
	\centering
	\subfigure[Precision on Category]{\label{fig:substructure_precision_newyork}\includegraphics[width=4.35cm]{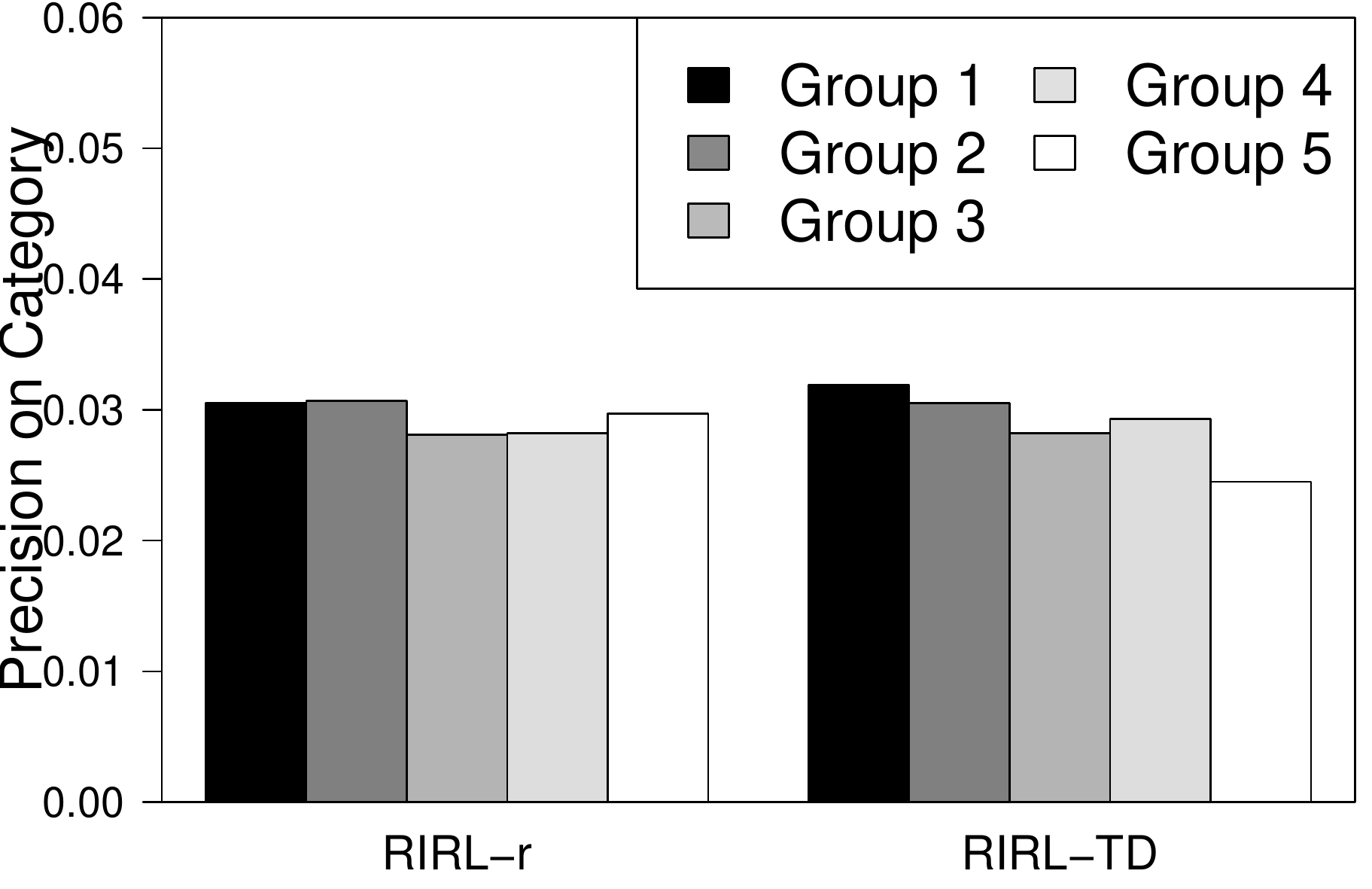}}
	\subfigure[Recall on Category]{\label{fig:substructure_newprecision_newyork}\includegraphics[width=4.35cm]{{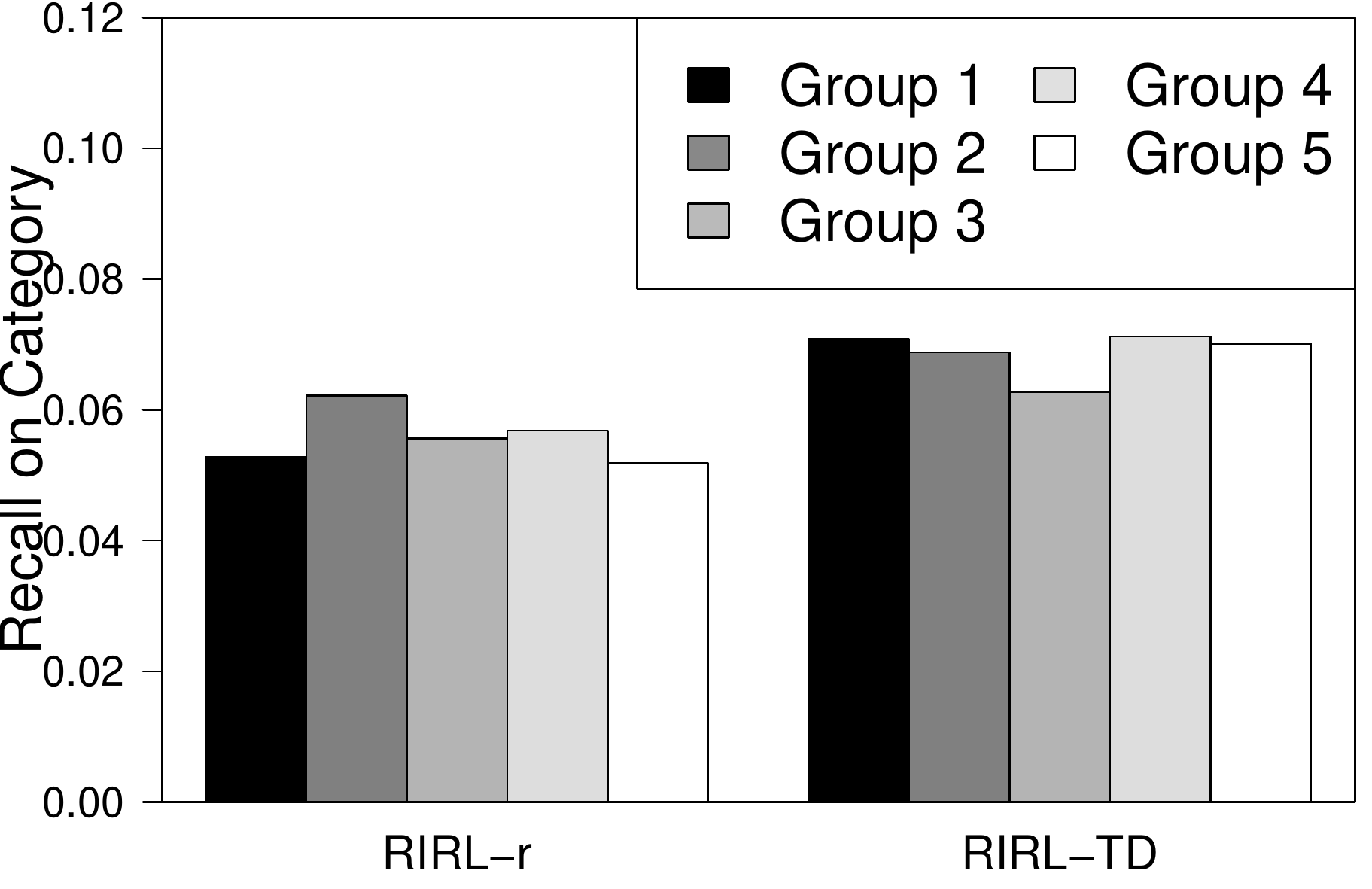}}}
	\subfigure[Average Similarity]{\label{fig:substructure_precision_tokyo}\includegraphics[width=4.35cm]{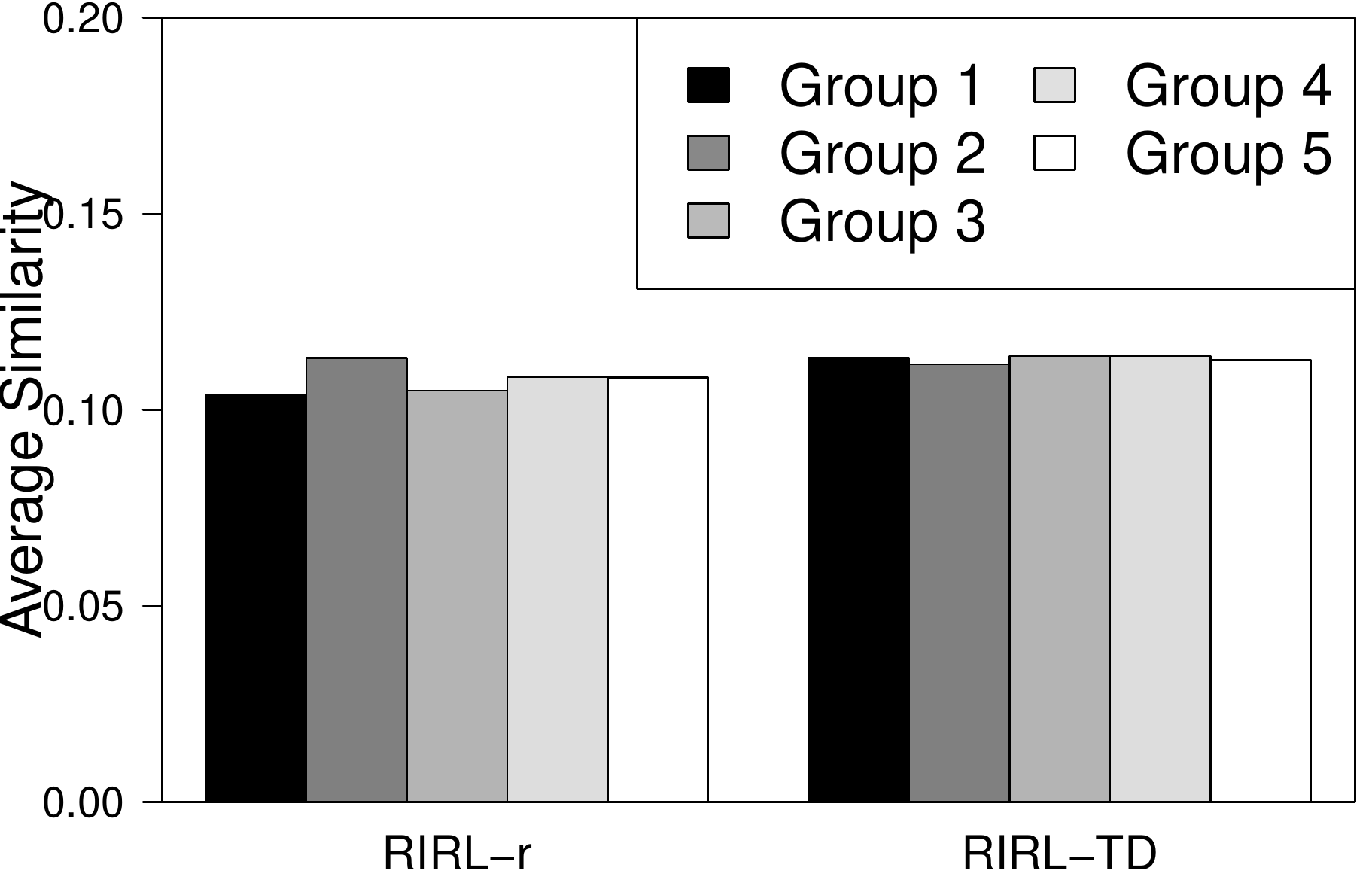}}
	\subfigure[Average Distance]{\label{fig:substructure_newprecision_tokyo}\includegraphics[width=4.35cm]{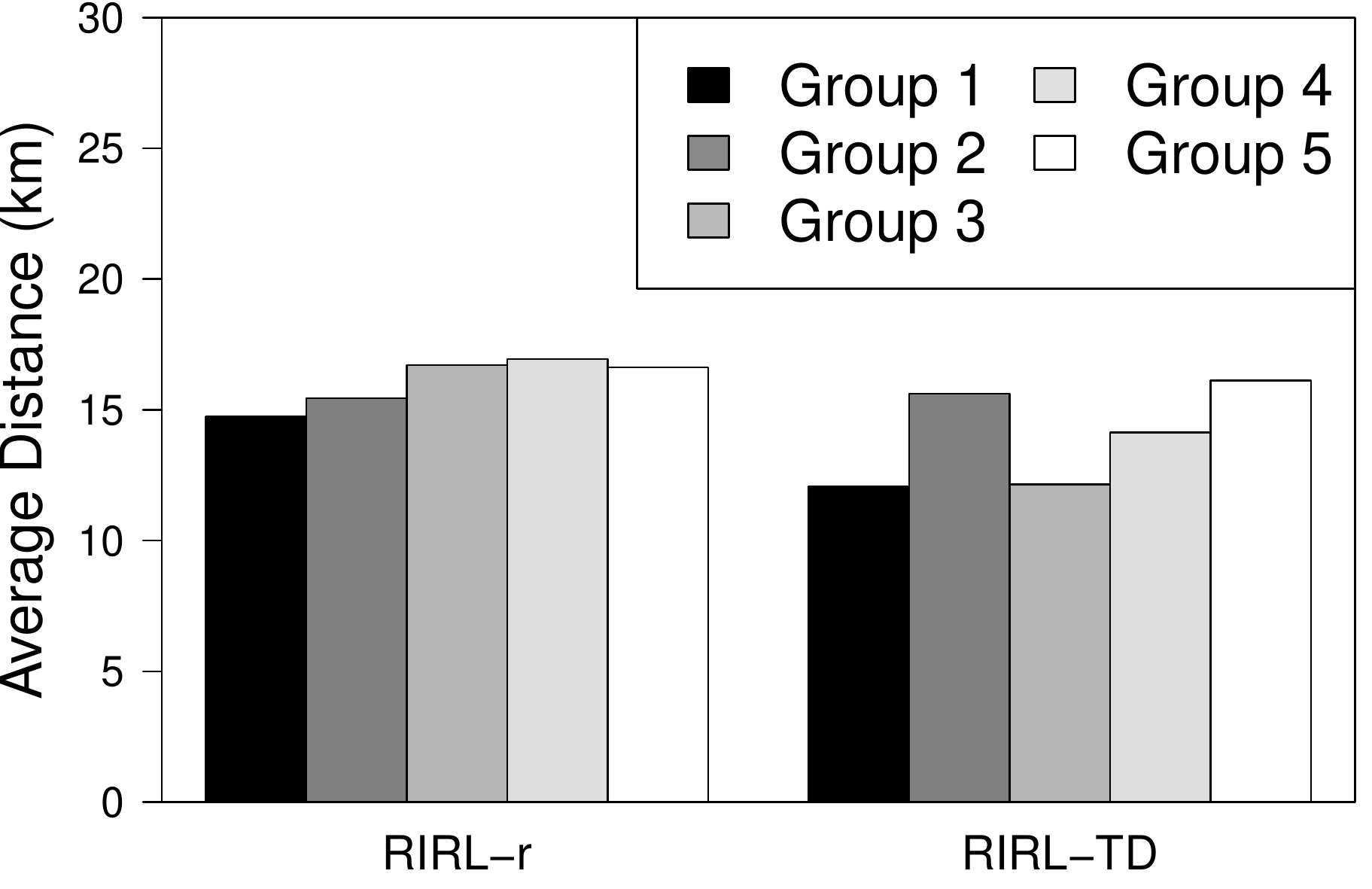}}
		\vspace{-0.45cm}
	\captionsetup{justification=centering}
	\caption{Robustness check {\it w.r.t.} New York dataset.}
		\vspace{-0.35cm}
	\label{fig:nyc_robust}
\end{figure*}

\begin{figure*}[!tb]
	\centering
	\subfigure[Precision on Category]{\label{fig:substructure_precision_newyork}\includegraphics[width=4.35cm]{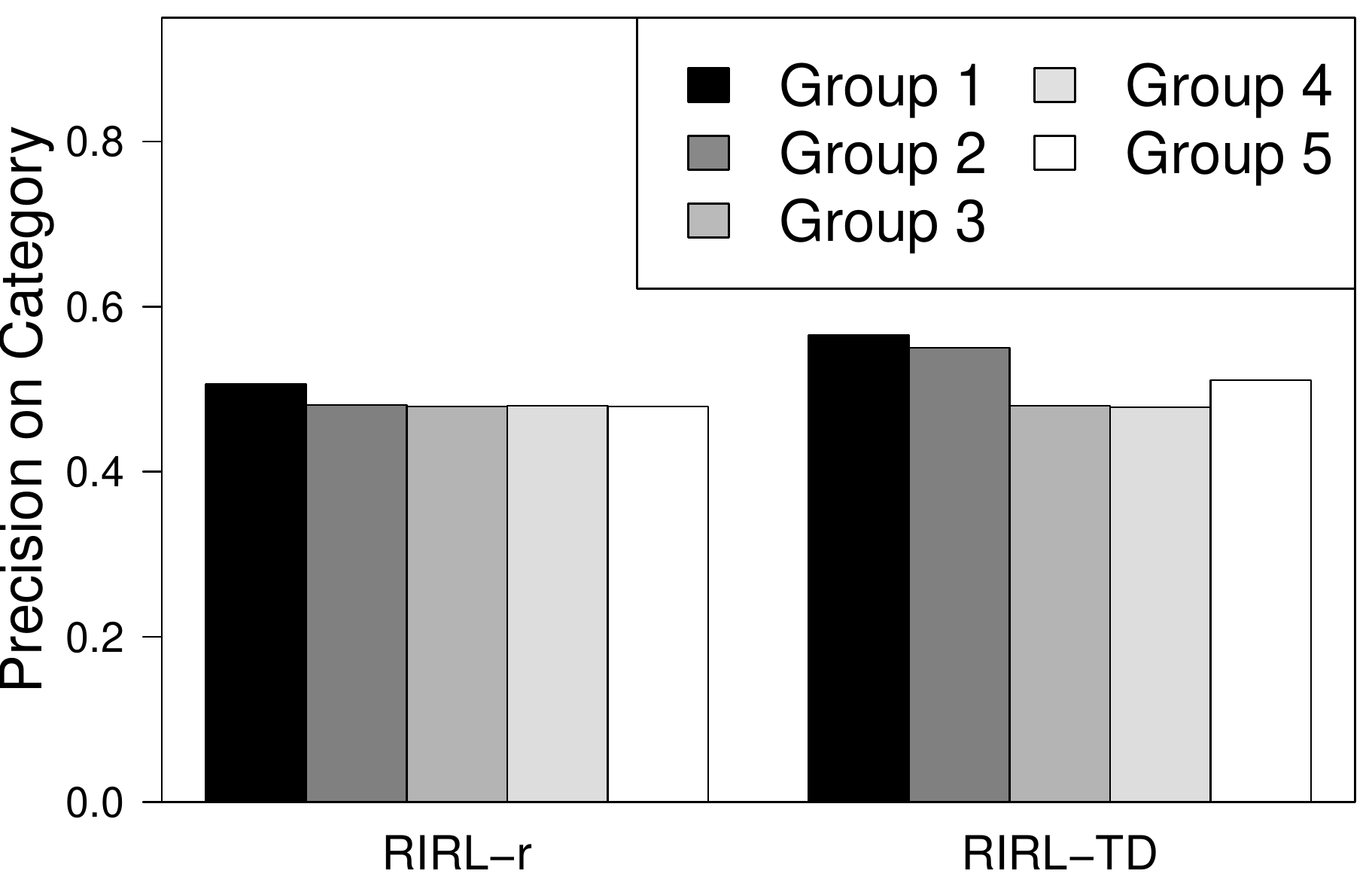}}
	\subfigure[Recall on Category]{\label{fig:substructure_newprecision_newyork}\includegraphics[width=4.35cm]{{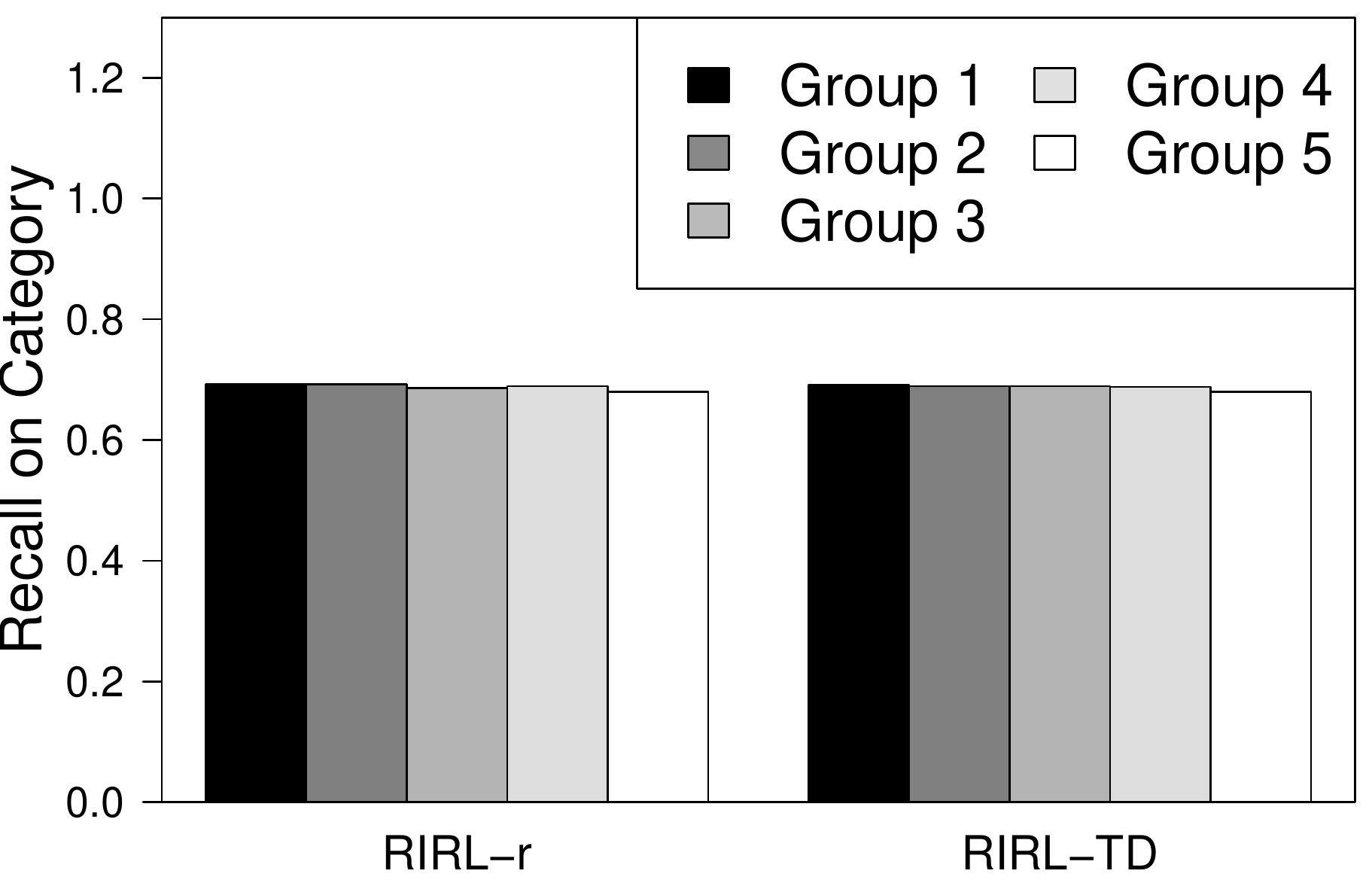}}}
	\subfigure[Average Similarity]{\label{fig:substructure_precision_tokyo}\includegraphics[width=4.35cm]{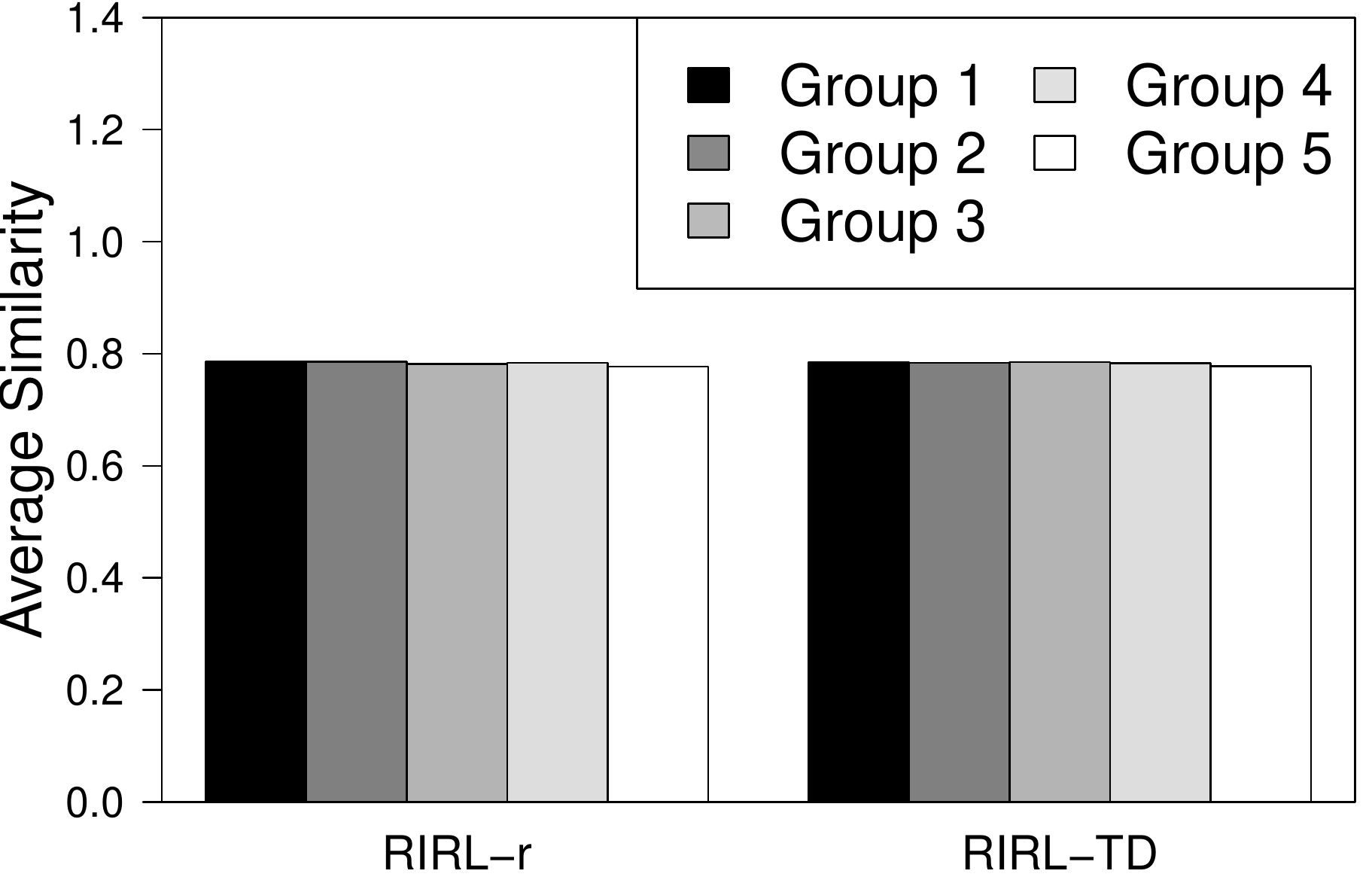}}
	\subfigure[Average Distance]{\label{fig:substructure_newprecision_tokyo}\includegraphics[width=4.35cm]{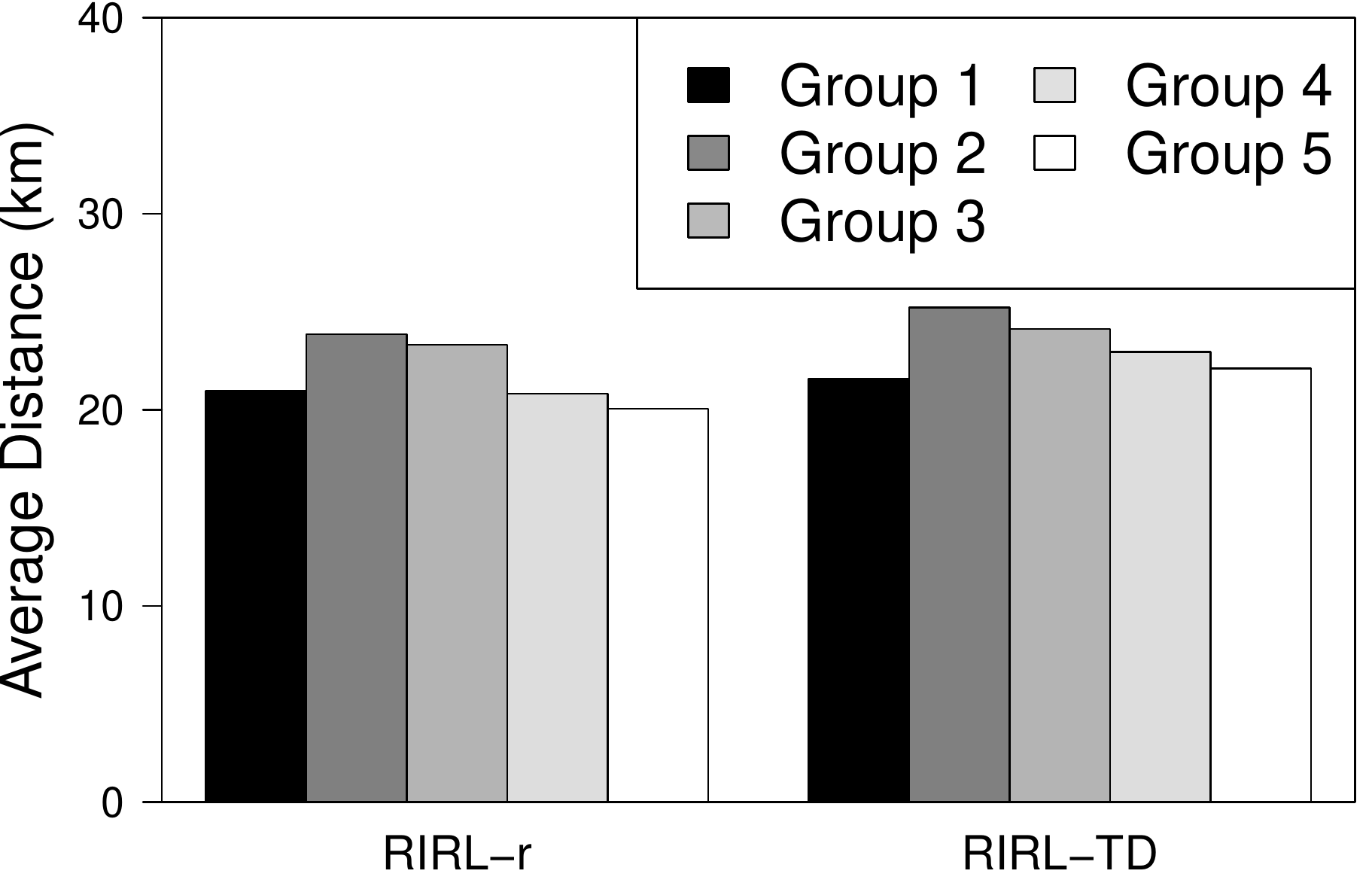}}
		\vspace{-0.45cm}
	\captionsetup{justification=centering}
	\caption{Robustness check {\it w.r.t.} Beijing dataset.}
		\vspace{-0.55cm}
	\label{fig:bj_robust}
\end{figure*}
Meanwhile, it indicates that RIRL retains the most representative user characteristics in user profiles.
The imitation module can predict more accurately based on the effective user profiles.
In addition, compared with ``RIRL-r'' and ``RIRL-TD'', we find that ``RIRL-TD'' defeats ``RIRL-r'' in terms of all evaluation metrics.
This phenomenon shows the TD-based sampling strategy is more superior than the reward-based sampling strategy in this profiling task.



\subsection{Effectiveness of the Adversarial Training and Gated Mechanism}
``RIRL'' is an enhanced version of ``IMUP'' \cite{wang2020incremental}. 
Comparing to ``IMUP'', ``RIRL'' proposes a new adversarial training strategy for robust user profiles, and exploits the gated mechanism for modeling temporal dependency of user profiles.
Figure \ref{fig:nyc_overall} and Figure \ref{fig:bj_overall} show that ``RIRL'' surpasses ``IMUP''  in terms of all evaluation metrics under the same sampling strategy.
This result verifies the effectiveness of the adversarial training and gated mechanism.
A possible explanation is that the adversarial training strategy keeps the training process of the representation module stable, which further enhances the capability of the imitation module.
In addition, the gated mechanism is helpful to discard trivial information in the old profile, and incorporate important new user interests to update new user profiles.

\vspace{-0.2cm}
\subsection{Robustness Check}
To evaluate the robustness of RIRL, we evenly divide the human mobility event sequence into five groups.
Then, we conduct experiments over these groups.
Figure \ref{fig:nyc_robust} and Figure \ref{fig:bj_robust}
show the experimental results.
We can find that the performance of RIRL is relatively stable over the five groups. 
Such observation indicates that the proposed adversarial training strategy mitigates the random issue transmitted from the imitation module. 
Due to the exploration in policy learning, the imitation module may generate biased actions (POI visit) accidentally. 
Then, such randomness is fed back to the representation module to jeopardize the quality of user profiles. 
Our proposed adversarial training strategy guides user profiles to update in a high-reward direction. 
Consequently, the incrementally improved the user profiles guarantee the stable performance of the imitation module through the closed-loop learning procedure.

%% file: related.tex
\vspace{-0.1cm}
\section{Related Works}
\noindent{\bf Static Mobile User Profiling}
Our work is mainly related to mobile user profiling.
User profiling methods can be classified into two categories:
(1) explicit feature extraction, which uses explicit characteristics such as age, sex, etc to represent user profile. 
(2) model-based approach, which
preserves user pattern in user profile by models.
Our RIRL model belongs to model-based approach.
Model-based approaches are popular.
For example, in \cite{griesner2015poi}, the researchers exploit factorization-based approaches to construct user representation by integrating spatio-temporal influence of human mobility.
In \cite{wang2019adversarial}, adversarial learning  is used to generate user profiling by focusing on the substructure of user mobility patterns.
In \cite{pan2020learning}, they use deep auto encoder model to generate user representation by minimizing reconstruction loss \cite{pan2020learning}.

\noindent{\bf Reinforcement Learning for Incremental User Profiling.}
Our work utilizes reinforcement learning to learn user profiling incrementally.
Researchers always regard online user behavior ({\it e.g.}, clicks, reviews, purchases) as a sequential decision-making process when reinforcement learning is used to obtain user profiling \cite{zhang2019deep}. 
For instance, 
Zhao {\it et al.} utilize reinforcement learning to provide feedback that captures the newest user interest for the recommendation system in real-time ~\cite{zhao2018deep}.
Wang {\it et al.} incorporate spatial knowledge graph and reinforcement learning to capture the user characteristics incrementally \cite{wang2020incremental}.
Our work is similar to \cite{wang2020incremental}, but  we explain the user profiling framework from an adversarial learning perspective and propose an adversarial training strategy and gated mechanism to improve the performance of the user profiling framework.


%% file: conclusion.tex
\section{Conclusion Remarks}
In this paper, we propose a new user profiling framework (RIRL) that integrates representation and imitation in reinforcement learning to learn effective user profiles incrementally.
RIRL is built on an imitation-based learning criterion that the user profiles are optimal once an agent can perfectly imitate the user mobility behavior. 
Specifically, RIRL includes the representation module and imitation module, where the representation module generates user profiles, and the imitation module reproduces the user mobility behavior based on the user profiles. 
The imitation performance is evaluated as the feedback to update user profile in real-time.
In order to keep the learning procedure of the representation module robust and retains the most representative user characteristics in user profiles, we propose an adversarial training strategy for training RIRL and introduce a gated mechanism to update user profiles.
From experimental results, we find that RIRL outperforms other baselines in terms of all evaluation metrics, which supports the effectiveness of the hypothesis of RIRL.